\documentclass[11pt,letterpaper]{article}
\pdfoutput=1
\usepackage[letterpaper]{geometry}
\usepackage{acl2012}
\usepackage{times}
\usepackage{latexsym}
\usepackage{enumitem}
\usepackage{graphicx}
\usepackage[colorlinks=true, urlcolor=black, linkcolor=blue, citecolor = black]{hyperref}

\usepackage[latin9]{inputenc}

\usepackage{graphicx} 
\usepackage{subfigure} 
\usepackage{tikz}

\usepackage{amsmath} 
\usepackage{amssymb} 
\usepackage{amsthm}  
\usepackage{relsize}

\usepackage{booktabs}

\usepackage{multirow}

\usepackage{color}

\usepackage{xspace} 

\usepackage{comment}

\usepackage{bm}
\usepackage{algorithm}

\def\trans{^{\sf T}}              
\newcommand{\dif}{\mathrm{d}}     

\def\bX{\mathbf{X}}

\def\bl{\mathbf{l}}
\def\br{\mathbf{r}}
\def\by{\mathbf{y}}

\def\bI{\mathbf{I}}

\def\bI{\mathbf{I}}

\def\bS{\mathbf{S}}

\def\bu{\mathbf{u}}

\def\balpha{\boldsymbol\alpha}

\def\bdelta{\boldsymbol\delta}
\def\bgamma{\boldsymbol\gamma}

\def\bGamma{\boldsymbol\Gamma}
\def\bpi{\boldsymbol\pi}
\def\bEta{\boldsymbol\eta}

\definecolor{mygreen}{rgb}{0,0.3,0}

\def\ourModel{\textsl{Semiparametric}\xspace} 
\def\gammaModel{\textsl{Landwehr et al.}\xspace} 
\def\gammaModelT{\textsl{Landwehr et al. (T)}\xspace} 
\def\gammaModelTA{\textsl{Landwehr et al. (TA)}\xspace} 
\def\hollandWeighted{\textsl{Holland \& K. (weighted)}\xspace} 
\def\hollandUnweighted{\textsl{Holland \& K. (unweighted)}\xspace}

\title{A Semiparametric Model for Bayesian Reader Identification} 

\author{
Ahmed Abdelwahab$^1$ \and Reinhold Kliegl$^2$ \and Niels Landwehr$^1$\\
$^1$ Department of Computer Science, Universit\"at Potsdam\\
August-Bebel-Straße 89, 14482 Potsdam, Germany\\
\{ahmed.abdelwahab, niels.landwehr\}@uni-potsdam.de\\
$^2$ Department of Psychology, Universit\"at Potsdam\\
Karl-Liebknecht-Straße 24/25, 14476 Potsdam OT/Golm\\
kliegl@uni-potsdam.de\\
}
\date{}

\begin{document}
\maketitle

\begin{abstract}

We study the problem of identifying individuals based on their characteristic gaze patterns during reading of arbitrary text.
The motivation for this problem is an unobtrusive biometric setting in which a user
is observed during access to a document, but no specific challenge protocol requiring the user's time and attention
 is carried out.
Existing models of individual differences in gaze control during reading are either based on simple aggregate features of 
eye movements, or rely on parametric density models to describe, for instance, saccade amplitudes or word fixation durations.
We develop flexible semiparametric models of eye movements during reading in which densities are inferred under a Gaussian process prior
centered at a parametric distribution family that is expected to approximate the true distribution well.
An empirical study on reading data from 251 individuals shows significant improvements over the state of the art. 
\end{abstract}

\section{Introduction}

Eye-movement patterns during skilled reading consist of brief fixations of individual words in a text that are interleaved with 
quick eye movements called \emph{saccades} that change the point of fixation to another word.
Eye movements are driven both by low-level visual cues and high-level linguistic and cognitive processes related to text understanding; 
as a reflection of the interplay between vision, cognition, and motor control during reading they are frequently studied in cognitive psychology~\cite{kliegl2006tracking,rayner1998review}.
Computational models~\cite{engbert2005swift,reichle1998toward} as well as models based on machine learning~\cite{matties2013with,hara2012predicting} have been developed to study how gaze 
patterns arise based on text content and structure, facilitating the understanding of human reading processes. 

A central observation in these and earlier psychological studies~\cite{huey1908reading,dixon1951studies} is that eye movement patterns strongly differ between individuals.
Holland et al.~\shortcite{holland2012biometric} and Landwehr et al.~\shortcite{landwehr2014model} have developed models of individual differences in eye movement patterns during reading,
and studied these models in a biometric problem setting where an individual has to be identified based on observing her eye movement patterns while reading
arbitrary text. 
Using eye movements during reading as a biometric feature has the advantage that it suffices to observe a user during a routine access to a device or document,
without requiring the user to react to a specific challenge protocol.
If the observed eye movement sequence is unlikely to be generated by an authorized individual, access can be
terminated or an additional verification requested. 
This is in contrast to approaches where biometric identification is based on eye movements in response to an artificial visual stimulus,
for example a moving~\cite{kasprowski2004eye,komogortsev2010biometric,rigas2012human,zhang2012on} or fixed~\cite{bednarik2005eye}  dot on a computer screen, or a specific image stimulus~\cite{rigas2012biometric}.

The model studied by Holland \& Komogortsev~\shortcite{holland2012biometric} uses aggregate features (such as average fixation duration) of the observed
eye movements. Landwehr et al.~\shortcite{landwehr2014model} showed that readers can be identified more accurately with a model that captures aspects of individual-specific distributions over eye movements,
such as the distribution over fixation durations or saccade amplitudes for word refixations, regressions, or next-word movements. 
Some of these distributions need to be estimated from very few observations;
a key challenge is thus to design models that are flexible enough to capture characteristic differences
between readers yet robust to sparse data. Landwehr et al.~\shortcite{landwehr2014model} used a fully parametric approach where all densities are assumed to be in the gamma family;
gamma distributions were shown to approximate the true distribution of interest well for most cases (see Figure~\ref{fig:distributions}). 
This model is robust to sparse data, but might not be flexible enough to capture all differences between readers.

The model we study in this paper follows ideas developed
by Landwehr et al.~\shortcite{landwehr2014model}, but employs more flexible semiparametric density models. 
Specifically, we place a Gaussian process prior over densities that concentrates probability mass on densities that are close to the gamma family.
Given data, a posterior distribution over densities is derived. If data is sparse, the posterior will still be sharply peaked around distributions in the gamma family, reducing the 
effective capacity of the model and minimizing overfitting. However, given enough evidence in the data, the model will also deviate from the gamma-centered prior and represent any density function. 
Integrating over the space of densities weighted by the posterior yields a marginal likelihood for novel observations from which predictions are inferred. 
We empirically study this model in the same setting as studied by Landwehr et al.~\shortcite{landwehr2014model}, but using an order of magnitude more individuals.
Identification error is reduced by more than a factor of three compared to the state of the art.

The rest of the paper is organized as follows. After defining the problem setting in Section~\ref{sec:problem_setting}, Section~\ref{sec:model} presents the semiparametric probabilistic model. 
Section~\ref{sec:inference} discusses inference, Section~\ref{sec:experiments} presents an empirical study on reader identification. 

\section{Problem Setting}
\label{sec:problem_setting}

Assume $R$ different readers, indexed by $r \in \{1,\ldots,R\}$, and let \mbox{$\mathcal{X}=\{\bX_1,\ldots,\bX_{n}\}$}
denote a set of texts.
Each \mbox{$r \in \mathcal{R}$} generates a set of eye movement patterns \mbox{$\mathcal{S}^{(r)} = \{\bS^{(r)}_1,\dots,\bS^{(r)}_{n}\}$}
on $\mathcal{X}$, by
\begin{equation*}
\bS^{(r)}_i \sim p(\bS|\bX_i,r,\bGamma)
\end{equation*}
where $p(\bS|\bX_i,r,\bGamma)$ is a reader-specific distribution over eye movement patterns given a text $\bX_i$.
Here, $r$ is a variable indicating the reader generating the sequence, and $\bGamma$ is a true but unknown model that defines all reader-specific distributions. 
We assume that $\bGamma$ can be broken down into reader-specific models, $\bGamma = (\bgamma_1,\ldots,\bgamma_R)$,
such that the distribution
\begin{equation}
p(\bS|\bX_i,r,\bGamma) = p(\bS|\bX_i,\bgamma_r)\label{eq:reader_specific_model}
\end{equation}
is defined by the partial model $\bgamma_r$. 
We aggregate the observations of all readers on the training data into a variable \mbox{$\mathcal{S}^{(1:R)} = (\mathcal{S}^{(1)},\ldots,\mathcal{S}^{(R)})$}.

We follow a Bayesian approach, defining a prior $p(\bGamma)$ over the joint model that 
factorizes into priors over reader-specific models, $p(\bGamma) = \prod_{r=1}^R p(\bgamma_r)$.
At test time, we observe novel eye movement patterns $\bar{\mathcal{S}} = \{\bar{\bS}_1,\dots,\bar{\bS}_{m}\}$ on a novel set of texts \mbox{$\bar{\mathcal{X}}=\{\bar{\bX}_1,\ldots,\bar{\bX}_{m}\}$} 
generated by an unknown reader $r\in\mathcal{R}$. We assume a uniform prior over readers, that is, each $r\in\mathcal{R}$ is equally
likely to be observed at test time.
The goal is to infer the most likely reader to have generated the 
novel eye movement patterns. In a Bayesian setting, this means inferring the most likely reader given 
the training observations ($\mathcal{X},\mathcal{S}^{(1:R)}$) and test observation ($\bar{\mathcal{X}},\bar{\mathcal{S}}$):
\begin{equation}
\label{eq:goal}
r_* = \arg\max_{r \in \mathcal{R}} p(r|\bar{\mathcal{X}},\bar{\mathcal{S}},\mathcal{X},\mathcal{S}^{(1:R)}).
\end{equation}
We can rewrite Equation~\ref{eq:goal} to
\begin{align}
r_* &= \arg\max_{r \in \mathcal{R}} p(\bar{\mathcal{S}}|r,\bar{\mathcal{X}},\mathcal{X},\mathcal{S}^{(1:R)}) \label{eq:goal_uniform_r}\\
&= \arg\max_{r \in \mathcal{R}}\hspace{-0.3mm}  \int \hspace{-0.3mm} p(\bar{\mathcal{S}}|r,\bar{\mathcal{X}},\bGamma)p(\bGamma|\mathcal{X},\mathcal{S}^{(1:R)}) \dif \bGamma\notag\\
&= \arg\max_{r \in \mathcal{R}}\hspace{-0.3mm} \int \hspace{-0.3mm} p(\bar{\mathcal{S}}|\bar{\mathcal{X}},\bgamma_r)p(\bgamma_r|\mathcal{X},\mathcal{S}^{(r)}) \dif \bgamma_r\label{eq:goal_factorization}
\end{align}
where
\begin{align}
p(\bar{\mathcal{S}}|\bar{\mathcal{X}},\bgamma_r) &= \prod_{i=1}^m p(\bar{\bS}_i|\bar{\bX}_i,\bgamma_r)\label{eq:test_likelihood_reader}\\ 
p(\bgamma_r|\mathcal{X},\mathcal{S}^{(r)}) &\propto p(\bgamma_r) \prod_{i=1}^n p(\bS_i^{(r)}|\bX_i,\bgamma_r)\label{eq:posterior_reader}.
\end{align}
In Equation~\ref{eq:goal_uniform_r} we exploit that readers are uniformly chosen at test time, and in Equation~\ref{eq:goal_factorization} we exploit the
factorization $p(\bGamma) = \prod_{r=1}^R p(\bgamma_r)$ of the prior, which together with Equation~\ref{eq:reader_specific_model} entails a factorization 
$p(\bGamma|\mathcal{X},\mathcal{S}^{(1:R)}) = \prod_{r=1}^R p(\bgamma_r|\mathcal{X},\mathcal{S}^{(r)})$ of the posterior.
Note that Equation~\ref{eq:goal_factorization} states that at test time we predict the reader $r$ for which the marginal likelihood (that is,
after integrating out the reader-specific model $\gamma_r$) of the test observations is highest.
The next section discusses the reader-specific models $p(\bS|\bX,\bgamma_r)$ and prior distributions $p(\bgamma_r)$.

\section{Probabilistic Model}
\label{sec:model}

\begin{figure}[t]
	\begin{center}
		\includegraphics[width=0.47\linewidth,trim=25 170 25 170, clip]{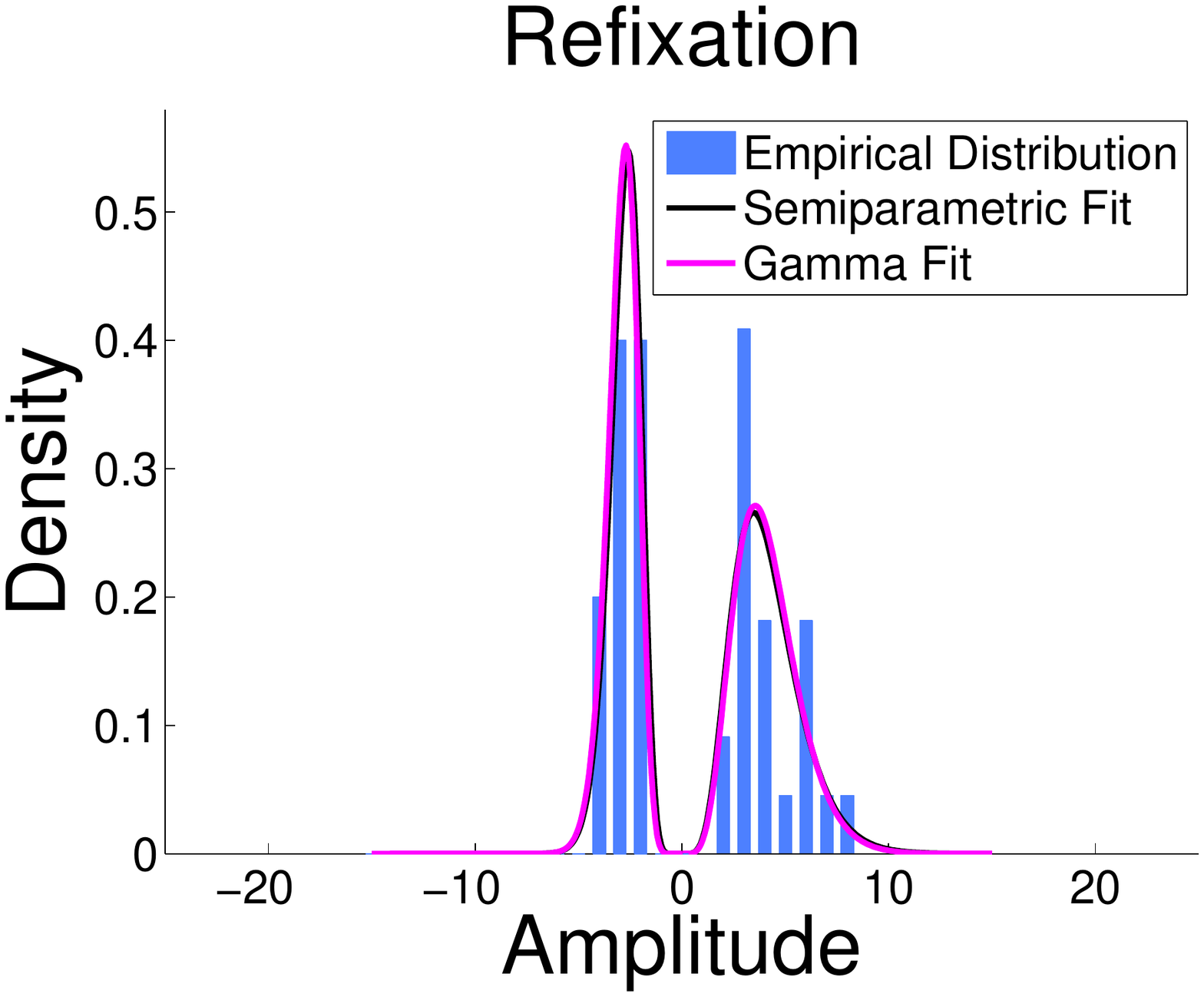}\hspace{3mm}		
		\includegraphics[width=0.47\linewidth,trim=25 170 25 170, clip]{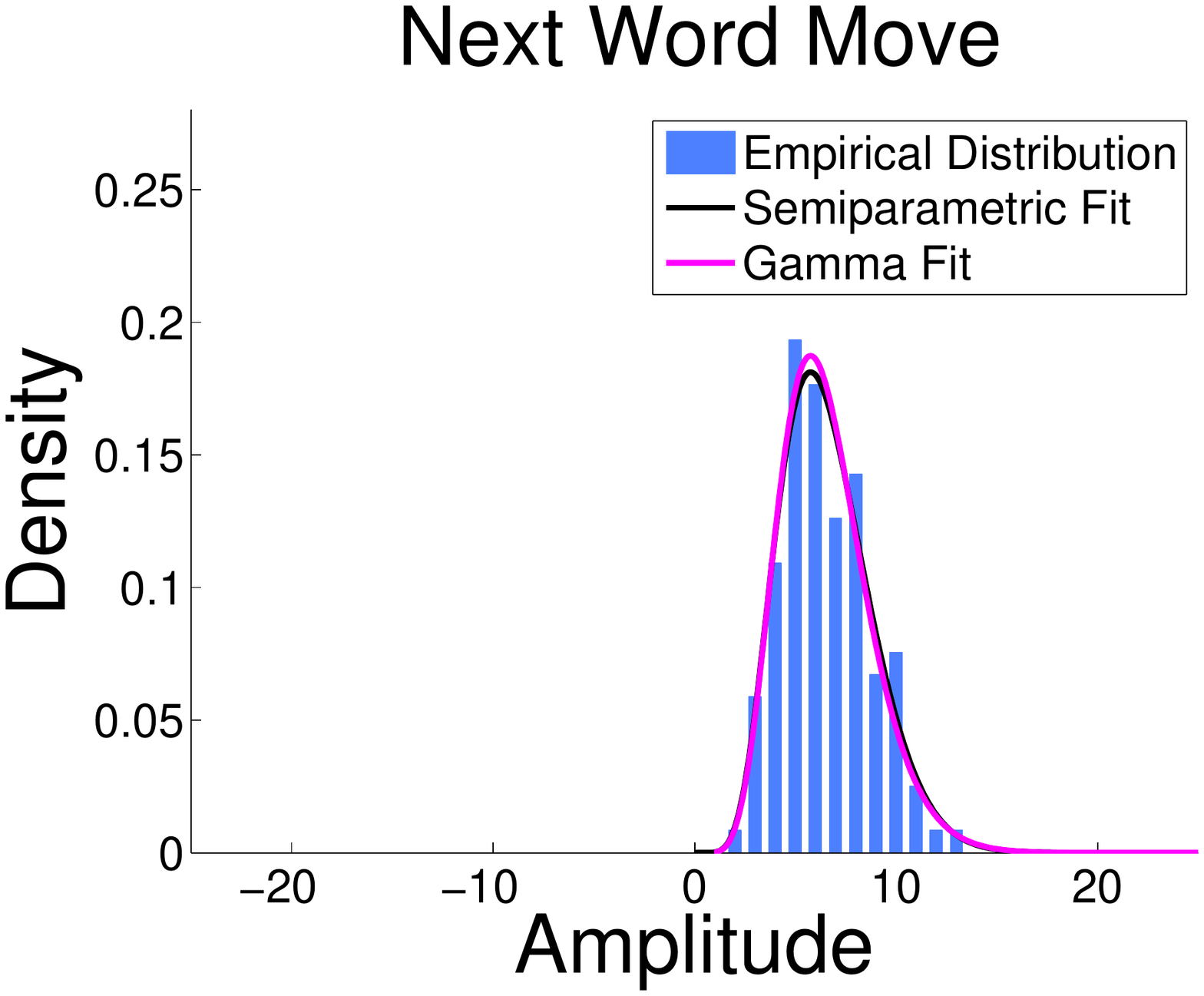}
		\vspace{2mm}
		\\
			\includegraphics[width=0.47\linewidth,trim=25 170 25 170, clip]{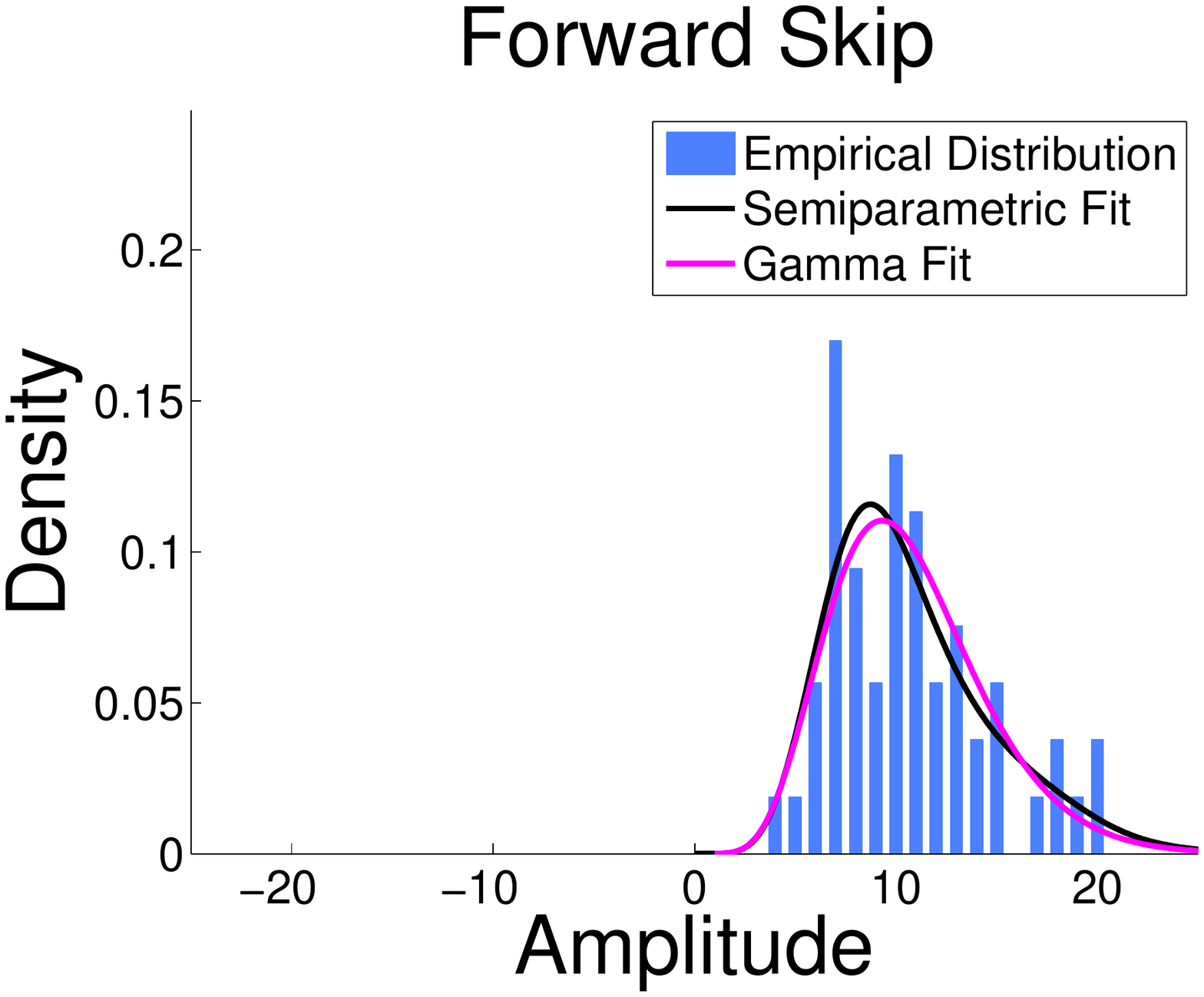}\hspace{3mm}
		\includegraphics[width=0.47\linewidth,trim=25 170 25 170, clip]{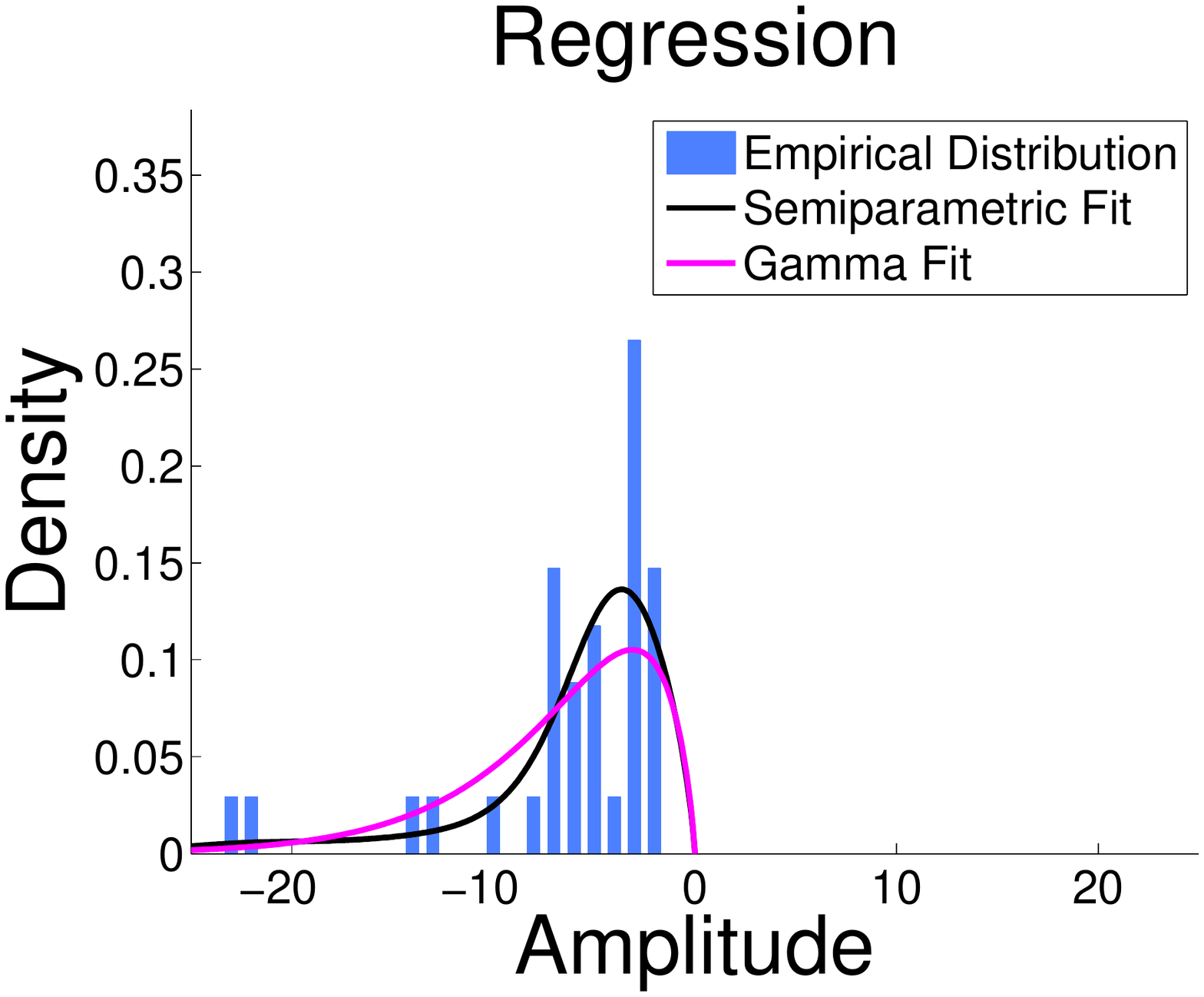}
		\caption{
		Empirical distributions of saccade amplitudes in training data for first individual, with fitted Gamma distributions and semiparametric distribution fits. 
		}
		\label{fig:distributions}
	\end{center}
\end{figure}

\begin{figure}[t]
	\begin{center}
		\includegraphics[width=0.75\linewidth,viewport=120 65 480 425, clip]{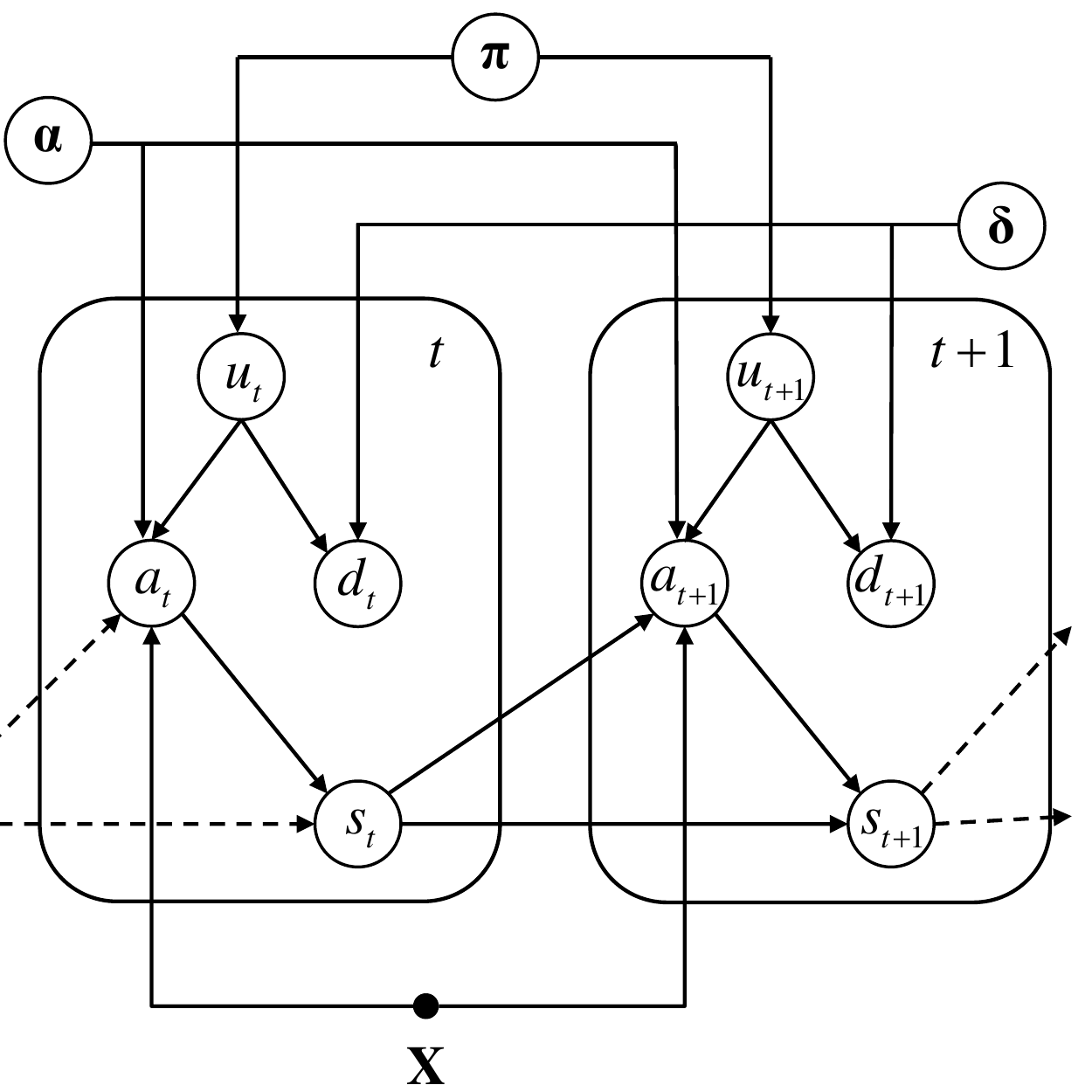}
		\caption{
		Plate notation of of a slice in the dynamic model. 
		}
		\label{fig:graphical_model}
	\end{center}
\end{figure}

The probabilistic model we employ follows the general structure proposed by Landwehr et al.~\shortcite{landwehr2014model}, but employs semiparametric 
density models and allows for fully Bayesian inference.
To reduce notational clutter, let $\bgamma \in \{\bgamma_1,\ldots,\bgamma_R\}$ denote a particular reader-specific model, and let $\bX \in \mathcal{X}$ denote a text.
An eye movement pattern is a sequence $\bS=\left((s_1,d_1),\dots,(s_T,d_T)\right)$ of gaze fixations, consisting of a fixation position $s_t$ (position in text that was fixated) and duration $d_t \in \mathbb{R}$ (length of fixation in milliseconds).
In our experiments, 
individual sentences are presented in a single line 
on screen, thus we only model a horizontal gaze position $s_t \in \mathbb{R}$. 
We model $p(\bS|\bX,\bgamma)$ 
as a dynamic process that successively generates fixation positions $s_t$ and durations $d_t$ in $\bS$, reflecting how a reader generates a sequence of saccades in response to a text stimulus $\bX$: 
\begin{align*}
p(\bS|&\bX, \bgamma)\notag =p(s_1,d_1|\bX,\bgamma)\prod_{t=2}^{T} p(s_t,d_t|s_{t-1},\bX,\bgamma),
\end{align*}
where $p(s_t,d_t|s_{t-1},\bX,\bgamma)$ models the generation of the next fixation position and duration given the old fixation position $s_{t-1}$.
In the psychological literature, four different \emph{saccade types} are distinguished:
a reader can refixate the current word (\emph{refixation}), fixate the next word in the text (\emph{next word movement}),
move the fixation to a word after the next word, that is, skip one or more words (\emph{forward skip}), or regress to fixate a word occurring earlier in the 
text (\emph{regression}), see, e.g., Heister et al.~\shortcite{heister2012analysing}. 
We observe empirically that for each saccade type, there is a characteristic distribution over saccade amplitudes and fixation durations, and that 
both approximately follow gamma distributions---see Figure~\ref{fig:distributions}. 
We therefore model $p(s_t,d_t|s_{t-1},\bX,\bgamma)$ using a mixture over distributions for the four different saccade types.
At each time $t$, the model first draws a saccade type \mbox{$u_t\in \{1,2,3,4\}$}, and then draws a saccade amplitude $a_t$ and fixation duration $d_t$ from type-specific 
distributions $p(a|u_t,s_{t-1},\bX,\bgamma)$ and $p(d|u_t,\bgamma)$.
More formally, 
\begin{align}
u_t &\sim p(u|\bpi)\label{eq:draw_u}\\
a_t &\sim p(a|u_t,s_{t-1},\bX,\balpha) \label{eq:draw_a} \\
d_t &\sim p(d|u_t,\bdelta) \label{eq:draw_d},
\end{align}
where $\bgamma = (\bpi,\balpha,\bdelta)$ is decomposed into components $\bpi$, $\balpha$, and $\bdelta$. 
Afterwards, the model updates the fixation position according to \mbox{$s_t = s_{t-1} + a_t$}, concluding the definition of
$p(s_t,d_t|s_{t-1},\bX,\bgamma)$.
Figure~\ref{fig:graphical_model} shows a slice in the dynamical model. 

The distribution $p(u|\bpi)$ over saccade types (Equation~\ref{eq:draw_u}) is multinomial with parameter vector $\bpi \in \mathbb{R}^4$.  
The distributions over amplitudes and durations (Equations~\ref{eq:draw_a} and~\ref{eq:draw_d}) are modeled semiparametrically
as discussed in the following subsections.

\subsection{Model of Saccade Amplitudes}
\label{sec:model_amplitudes}

We first discuss the amplitude model $p(a|u_t,s_{t-1},\bX,\balpha)$ (Equation~\ref{eq:draw_a}). We first define a distribution 
$p(a|u_t,\balpha)$ over amplitudes for saccade type $u_t$, and subsequently
discuss conditioning on the text $\bX$ and old fixation position $s_{t-1}$, leading to $p(a|u_t,s_{t-1},\bX,\balpha)$.
We define
\begin{align}
p(a|u_t=1,\balpha) = 
\begin{cases} \mu \alpha_1(a) \hspace{-2mm}&: a>0 \\
(1-\mu) \bar{\alpha}_1(-a) \hspace{-2mm}&: a \leq 0\label{eq:a_given_u1}
\end{cases} 
\end{align}
where $\mu$ is a mixture weight and $\alpha_1$, $\bar{\alpha}_1$ are densities defining the distribution over positive and negative amplitudes for the saccade type
\emph{refixation}, and
\begin{align}
p(a|u_t = 2,\balpha) &= \alpha_{2}(a) \label{eq:a_given_u2}\\
p(a|u_t = 3,\balpha) &= \alpha_{3}(a) \label{eq:a_given_u3}\\
p(a|u_t = 4,\balpha) &= \alpha_{4}(-a)\label{eq:a_given_u4}
\end{align}
where $\alpha_2(a)$, $\alpha_3(a)$, and $\alpha_4(a)$ are densities defining the distribution over amplitudes for the remaining saccade types.
Finally, the distribution 
\begin{equation}
p(s_1|\bX,\balpha) = \alpha_0(s_1)\label{eq:initial_amplitude}
\end{equation}
over the initial fixation position is given by another density function $\alpha_0$.
The variables $\mu,\alpha_0,\alpha_1$, $\bar{\alpha}_1$, $\alpha_2$, $\alpha_3$, and $\alpha_4$ are aggregated into model component $\balpha$.
For resolving the most likely reader at test time (Equation~\ref{eq:goal_factorization}), densities in $\balpha$ will be integrated
out under a prior based on Gaussian processes (Section~\ref{sec:priors}) using MCMC inference (Section~\ref{sec:inference}).

Given the old fixation position $s_{t-1}$, the text $\bX$, and the chosen saccade type $u_t$, the amplitude is constrained to fall within a specific interval.
For instance, for a refixation the amplitude has to be chosen such that the novel fixation position lies within the beginning and the end of the currently
fixated word; a regression implies an amplitude that is negative and makes the novel fixation position lie before the beginning of the currently 
fixated word.
These constraints imposed by the text structure define the conditional distribution $p(a|u_t,s_{t-1},\bX,\balpha)$.
More formally, $p(a|u_t,s_{t-1},\bX,\balpha)$ is the distribution $p(a|u_t,\balpha)$ conditioned on $a \in [l,r]$, that is, 
\begin{equation*}
p(a|u_t,s_{t-1},\bX,\balpha) = p(a|a\in[l,r],u_t,\balpha),
\end{equation*} 
where $l$ and $r$ are the minimum and maximum amplitude consistent with the constraints. 
Recall that for a distribution over a continuous variable $x$ given by density $\alpha(x)$, the distribution over $x$ conditioned on \mbox{$x \in [l,r]$} 
is given by the truncated density 
\begin{equation}
\alpha(x|x\in [l,r])=
\begin{cases} \frac{\alpha(x)}{\int_l^r \alpha(\bar{x})\dif \bar{x}}\hspace{-2mm}&: x \in [l,r]\\
0\hspace{-2mm}&:x \notin [l,r]. \end{cases} \label{eq:truncation}
\end{equation}
We derive $p(a|u_t,s_{t-1},\bX,\balpha)$ by truncating the distributions given by Equations~\ref{eq:a_given_u1} to~\ref{eq:a_given_u4} 
to the minimum and maximum amplitude consistent with the current fixation position $s_{t-1}$ and text $\bX$.
Let $w^{\circ}_l$ ($w^{\circ}_r$) denote the position of the left-most (right-most) character of the currently fixated word, and let $w^+_l,w^+_r$
denote these positions for the next word in $\bX$.  
Let furthermore \mbox{$l^{\circ} = w^{\circ}_l - s_{t-1}$}, $r^{\circ} = w^{\circ}_r - s_{t-1}$, \mbox{$l^+ = w^+_l-s_{t-1}$}, and $r^+ = w^+_r-s_{t-1}$.
Then
\begin{align}
&p(a|u_t=1,s_{t-1},\bX,\balpha) = \notag\\
&\hspace{8mm}\begin{cases} \mu \hspace{0.3mm} \alpha_1(a|a\in[0,r^{\circ}]) : a>0 \\
(1-\mu) \bar{\alpha}_1(-a|a\in[l^{\circ},0]) : a \leq 0 \label{eq:a_given_u1_truncated}
\end{cases}\\ 
&p(a|u_t \hspace{-0.4mm}=\hspace{-0.4mm} 2,s_{t-1},\bX,\balpha) \hspace{-0.4mm}= \hspace{-0.4mm}\alpha_{2}(a|a\hspace{-0.4mm} \in \hspace{-0.4mm} [l^+,r^+])\label{eq:a_given_u2_truncated} \\ 
&p(a|u_t \hspace{-0.4mm}=\hspace{-0.4mm} 3,s_{t-1},\bX,\balpha) \hspace{-0.4mm}= \hspace{-0.4mm}\alpha_{3}(a|a\hspace{-0.4mm} \in \hspace{-0.4mm} (r^+,\infty))\label{eq:a_given_u3_truncated}\\
&p(a|u_t \hspace{-0.4mm}=\hspace{-0.4mm} 4,s_{t-1},\bX,\balpha) \hspace{-0.4mm}= \hspace{-0.4mm}\alpha_{4}(-a|a\hspace{-0.4mm} \in \hspace{-0.4mm}(-\infty,l^{\circ}))\label{eq:a_given_u4_truncated} 
\end{align}
defines the appropriately truncated distributions.

\subsection{Model of Fixation Durations}
\label{sec:model_durations}

The model for fixation durations (Equation~\ref{eq:draw_d}) is similarly specified by saccade type-specific densities,
\begin{align}
p(d|u_t = u,\bdelta) = \delta_{u}(d) \text{\ \ \  for } u \in \{1,2,3,4\} \label{eq:d_given_u}
\end{align}
and a density for the initial fixation durations
\begin{equation}
p(d_1|\bX,\bdelta) = \delta_0(d_1)\label{eq:initial_duration}
\end{equation}
where $\delta_0,...,\delta_4$ are aggregated into model component $\bdelta$. 
Unlike saccade amplitude, the fixation duration is not constrained by the text structure and accordingly densities are not truncated. 
This concludes the definition of the model $p(\bS|\bX,\bgamma)$.

\subsection{Prior Distributions}
\label{sec:priors}

The prior distribution over the entire model $\bgamma$ factorizes over the model components as 
\begin{align}
&p(\bgamma|\lambda,\rho,\kappa) = \label{eq:complete_prior}\\
&\hspace{10mm}p(\bpi|\lambda)p(\mu|\rho)p(\bar{\alpha}_1|\kappa)\prod_{i=0}^4 p(\alpha_i|\kappa) \prod_{i=0}^4 p(\delta_i|\kappa)\notag
\end{align}
where $p(\bpi) = \text{Dir}(\bpi|\lambda)$ is a symmetric Dirichlet prior and $p(\mu) = \text{Beta}(\mu|\rho)$ is a Beta prior.
The key challenge is to develop appropriate priors for the densities defining saccade amplitude ($p(\bar{\alpha}_1|\kappa),p(\alpha_i|\kappa)$) and fixation duration ($p(\delta_i|\kappa)$) distributions.
Empirically, we observe that amplitude and duration distributions tend to be close to gamma distributions---see the example in Figure~\ref{fig:distributions}.

Our goal is to exploit the prior knowledge that distributions tend to be closely approximated by gamma distributions, but allow the model to deviate from the gamma assumption in case there is enough evidence in the data.
To this end, we define a prior over densities that concentrates probability mass around the gamma family.
For all densities \mbox{$f\in\{\bar{\alpha}_1,\alpha_0,...,\alpha_4,\delta_0,...,\delta_4\}$}, we employ identical prior distributions 
$p(f |\kappa)$. 
Intuitively, the prior is given by first drawing a density function from the gamma family and then drawing the final density from a 
Gaussian process (with covariance function $\kappa$) centered at this function. 
More formally, let
\begin{equation}
\mathcal{G}(x|\bEta) = \frac{\exp(\bEta\trans \bu(x))}{\int \exp(\bEta\trans \mathbf{u}(x'))\dif x'} \label{eq:gamma_exponential_family}
\end{equation}
denote the gamma distribution in exponential family form, with sufficient statistics $\bu(x)=(\log(x),x)\trans$ and parameters $\bEta = (\eta_1,\eta_2)$.
Let $p(\bEta)$ denote a prior over the gamma parameters, and define
\begin{equation}
p(f|\kappa) = \int p(\bEta) p(f|\bEta,\kappa) \dif \bEta \label{eq:prior_f}
\end{equation}
where $p(f|\bEta,\kappa)$ is given by drawing
\begin{equation}
g \sim \mathcal{GP}(0,\kappa) \label{eq:gp_prior}
\end{equation}
from a Gaussian process prior $\mathcal{GP}(0,\kappa)$ with mean zero and covariance function $\kappa$,
and letting
\begin{equation}
f(x) = \frac{\exp(\bEta\trans \bu(x) + g(x))}{\int \exp(\bEta\trans \bu(x')+g(x'))\dif x'}.\label{eq:prior_final_density}
\end{equation}
Note that decreasing the variance of the Gaussian process 
means regularizing $g(x)$ towards zero,
and therefore Equation~\ref{eq:prior_final_density} towards Equation~\ref{eq:gamma_exponential_family}.
This concludes the specification of the prior $p(\bgamma|\lambda,\rho,\kappa)$. 

The density model defined by Equations~\ref{eq:prior_f} to~\ref{eq:prior_final_density} draws on ideas from 
the large body of literature on GP-based density estimation, 
for example by Adams et al.~\shortcite{adams2009gaussian}, Leonard~\shortcite{leonard1978density}, or Tokdar et al.~\shortcite{tokdar2010bayesian},
and semiparametric density estimation, e.g. as discussed by Yang~\shortcite{yang2009penalized}, Lenk~\shortcite{lenk2003bayesian} or Hjort \& Glad~\shortcite{hjort1985nonparametric}.
However, note that existing density estimation approaches are not applicable off-the-shelf as in our domain distributions are truncated differently at each observation
due to constraints that arise from the way eye movements interact with the text structure (Equations \ref{eq:a_given_u1_truncated} to \ref{eq:a_given_u4_truncated}).

\section{Inference}
\label{sec:inference}

To solve Equation~\ref{eq:goal_factorization}, we need to integrate for each $r \in \mathcal{R}$ over the reader-specific model $\bgamma_r$.
To reduce notational clutter, let $\bgamma \in \{\bgamma_1,\ldots,\bgamma_R\}$ denote 
a reader-specific model, and let \mbox{$\mathcal{S} \in \{\mathcal{S}^{(1)},\ldots,\mathcal{S}^{(R)}\}$} denote the eye movement observations of that
reader on the training texts $\mathcal{X}$.
We approximate
\begin{align*}
&\int p(\bar{\mathcal{S}}|\bar{\mathcal{X}},\bgamma)p(\bgamma|\mathcal{X},\mathcal{S}) \dif \bgamma \approx  \frac{1}{K}\sum_{k=1}^K p(\bar{\mathcal{S}}|\bar{\mathcal{X}},\bgamma^{(k)})
\end{align*}
by a sample $\bgamma^{(1)},\ldots,\bgamma^{(K)}$ of models drawn by 
\begin{equation*}
\bgamma^{(k)} \sim p(\bgamma|\mathcal{X},\mathcal{S},\lambda,\rho,\kappa),
\end{equation*}
where $p(\bgamma|\mathcal{X},\mathcal{S},\lambda,\rho,\kappa)$ is the posterior as given by Equation~\ref{eq:posterior_reader} but with the dependence on the prior hyperparameters $\lambda,\rho,\kappa$ made explicit.
%
Note that with $\mathcal{X}$ and $\mathcal{S}$, all saccade types $u_t$ are observed. 
Together with the factorizing prior (Equation~\ref{eq:complete_prior}), this means that the posterior factorizes according to
\begin{align*}
&p(\bgamma|\mathcal{X},\mathcal{S},\lambda,\rho,\kappa) = p(\bpi|\mathcal{X},\mathcal{S},\lambda)p(\mu|\mathcal{X},\mathcal{S},\rho) \\
&\hspace{2mm}\cdot p(\bar{\alpha}_1|\mathcal{X},\mathcal{S},\kappa)\prod_{i=0}^4 p(\alpha_i|\mathcal{X},\mathcal{S},\kappa)\prod_{i=0}^4 p(\delta_i|\mathcal{X},\mathcal{S},\kappa)
\end{align*}
as is easily seen from the graphical model in Figure~\ref{fig:graphical_model}.
Obtaining samples $\bpi^{(k)} \sim p(\bpi|\mathcal{X},\mathcal{S})$ and $\mu^{(k)} \sim p(\mu|\mathcal{X},\mathcal{S})$ is straightforward because  
their prior distributions are conjugate to the likelihood terms. 
Let now \mbox{$f\in\{\bar{\alpha}_1,\alpha_0,...,\alpha_4,\delta_0,...,\delta_4\}$} denote a particular density in the model. 
The posterior $p(f|\mathcal{X},\mathcal{S},\kappa)$ is proportional to the prior $p(f|\kappa)$ (Equation~\ref{eq:prior_f}) 
multiplied by the likelihood of all observations that are generated by this density, that is, that are generated according to Equation~\ref{eq:initial_amplitude},~\ref{eq:a_given_u1_truncated},~\ref{eq:a_given_u2_truncated},~\ref{eq:a_given_u3_truncated},~\ref{eq:a_given_u4_truncated},~\ref{eq:d_given_u}, or~\ref{eq:initial_duration}. 
Let \mbox{$\by = (y_1,\ldots,y_{|\by|})\trans \in \mathbb{R}^{|\by|}$} denote the vector of all observations generated from density $f$, and let 
\mbox{$\bl = (l_1,\ldots,l_{|\bl|})\trans \in \mathbb{R}^{|\bl|}$}, $\br = (r_1,\ldots,r_{|\br|})\trans \in \mathbb{R}^{|\br|}$ denote the corresponding left and right
boundaries of the truncation intervals (again see Equations~\ref{eq:initial_amplitude} to~\ref{eq:initial_duration}), where for densities that are not truncated we take $l_i=0$ and $r_i=\infty$ throughout.  
Then the likelihood of the observations generated from $f$ is
\begin{equation}
p(\by|f,\bl,\br) = \prod_{i=1}^{|\by|}f(y_i|y_i\in [l_i,r_i])\label{eq:likelihood_density}
\end{equation}
and the posterior over $f$ is given by
\begin{equation}
p(f|\mathcal{X},\mathcal{S},\kappa) \propto p(f|\kappa)p(\by|f,\bl,\br). \label{eq:posterior_f}
\end{equation}
Note that $\by$, $\bl$ and $\br$ are observable from $\mathcal{X}$, $\mathcal{S}$. 

We obtain samples from the posterior given by Equation~\ref{eq:posterior_f}
from a Metropolis-Hastings sampler that explores the space of densities $f:\mathbb{R} \rightarrow \mathbb{R}$, generating density samples
$f^{(1)},...,f^{(K)}$.
A density $f$ is given by a combination of gamma parameters $\bEta\in \mathbb{R}^2$ and function $g:\mathbb{R}\rightarrow\mathbb{R}$; 
specifically, $f$ is obtained by multiplying the gamma distribution with parameters $\bEta$ by $\exp(g)$ and normalizing appropriately (Equation~\ref{eq:prior_final_density}).
During sampling, we explicitly represent a density sample $f^{(k)}$ by its gamma parameters $\bEta^{(k)}$ and function $g^{(k)}$.
The proposal distribution of the Metropolis-Hastings sampler is 
\begin{align*}
&q(\bEta^{(k+1)},g^{(k+1)}|\bEta^{(k)},g^{(k)}) = \\
&\hspace{30mm}p(g^{(k+1)}|\kappa)\mathcal{N}(\bEta^{(k+1)}|\bEta^{(k)},\sigma^2\bI)
\end{align*}
where $p(g^{(k+1)}|\kappa)$ is the probability of $g^{(k+1)}$ according to the GP prior $\mathcal{GP}(0,\kappa)$ (Equation~\ref{eq:gp_prior}), and
$\mathcal{N}(\bEta^{(k+1)}|\bEta^{(k)},\sigma^2\bI)$ is a symmetric proposal that randomly perturbs the old state $\bEta^{(k)}$ according to a Gaussian.
In every iteration $k$ a proposal $\bEta^{\star},g^{\star} \sim q(\bEta,g|\bEta^{(k)},g^{(k)})$
is drawn based on the old state $(\bEta^{(k)},g^{(k)})$. The acceptance probability is
$A(\bEta^{\star},g^{\star}|\bEta^{(k)},g^{(k)}) = \min(1,Q)$
with
\begin{align*}
&Q = \\
&\frac{
q(\bEta^{(k)},g^{(k)}|\bEta^{\star},g^{\star})p(\bEta^{\star})p(g^{\star}|\kappa) p(\by|f^{\star},\bl,\br)
}{
q(\bEta^{\star},g^{\star}|\bEta^{(k)},g^{(k)})p(\bEta^{(k)})p(g^{(k)}|\kappa),p(\by|f^{(k)},\bl,\br)
}.
\end{align*}
Here, $p(\bEta^{\star})$ is the prior probability of gamma parameters $\bEta^{\star}$ (Section~\ref{sec:priors})
and $p(\by|f^{\star},\bl,\br)$ is given by Equation~\ref{eq:likelihood_density} where $f^{\star}$ is obtained from $\bEta^{\star}$, $g^{\star}$
according to Equation~\ref{eq:prior_final_density}.

To compute the likelihood terms $p(\by|f^{(k)},\bl,\br)$ (Equation~\ref{eq:likelihood_density}) and also to compute the likelihood of test data under a model (Equation~\ref{eq:test_likelihood_reader}),
the density $f:\mathbb{R} \rightarrow \mathbb{R}$ needs to be evaluated.
According to Equation~\ref{eq:prior_final_density}, $f$ is represented by parameter vector $\bEta$ together with the nonparametric function $g:\mathbb{R}\rightarrow \mathbb{R}$. 
As usual when working with distributions over functions in a Gaussian process framework, the function $g$ only needs to be represented at those points
for which we need to evaluate it. Clearly, this includes all observations of saccade amplitudes and fixation durations observed
in the training and test set. 
However, we also need to evaluate the normalizer in Equation~\ref{eq:prior_final_density},
and (for $f\in\{\alpha_1,\bar{\alpha}_1,\alpha_2,\alpha_3,\alpha_4\}$) the additional normalizer required when truncating the distribution 
(see Equation~\ref{eq:truncation}). As these integrals are one-dimensional, they can be solved relatively accurately using numerical integration;
we use 2-point Newton-Cotes quadrature. Newton-Cotes integration requires the evaluation (and thus representation) of $g$ at an additional set of equally spaced supporting points.

When the set of test observations $\bar{\mathcal{S}},\bar{\mathcal{X}}$ is large, the need to evaluate $p(\bar{\mathcal{S}}|\bar{\mathcal{X}},\bgamma^{(k)})$ 
for all $\bgamma_k$ and all test observations leads to computational challenges. In our experiments, we use a heuristic to reduce computational load.
While generating samples, densities are only represented at the training observations and the supporting points needed for Newton-Cotes integration.
We then estimate the mean of the posterior by $\hat{\bgamma} = \frac{1}{K} \sum_{k=1}^K \bgamma^{(k)}$, and approximate
$\frac{1}{K}\sum_{k=1}^K p(\bar{\mathcal{S}}|\bar{\mathcal{X}},\bgamma^{(k)}) \approx p(\bar{\mathcal{S}}|\bar{\mathcal{X}},\hat{\bgamma})$.
To evaluate $p(\bar{\mathcal{S}}|\bar{\mathcal{X}},\hat{\bgamma})$, we infer the approximate value of the density $\hat{\bgamma}$ at a test observation
by linearly interpolating based on the available density values at the training observations and supporting points.

\section{Empirical Study}
\label{sec:experiments}

\begin{figure*}[t]
\centering
\includegraphics[width=0.99\linewidth,trim={0 15mm 0 5mm}, clip=true]{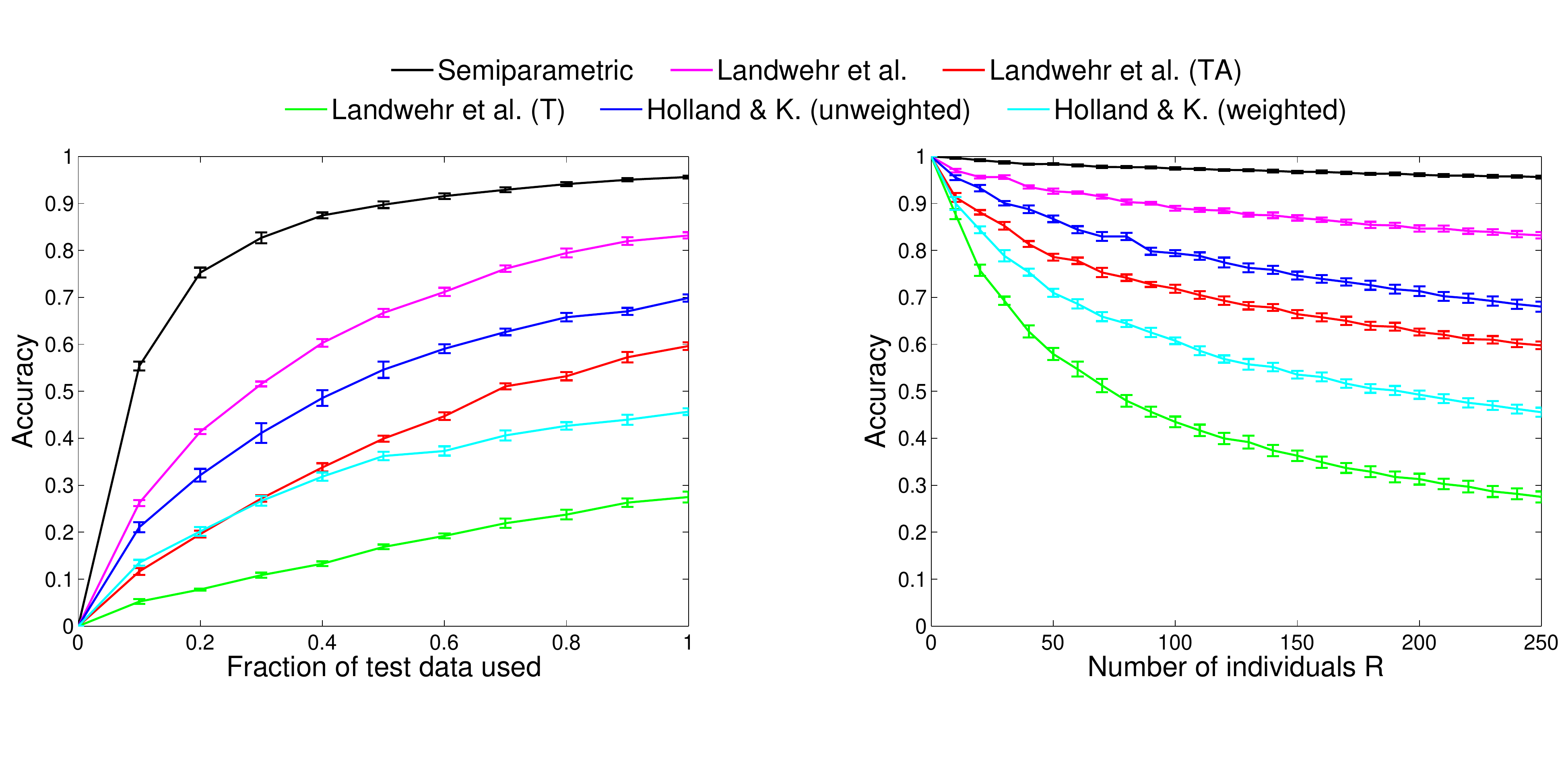}		
\caption{Multiclass accuracy over number of test observations (left) and number of individuals R (right) with standard errors. }
\label{fig:difftestindvmulti}
\end{figure*}

We conduct a large-scale study of biometric identification performance using the same setup as discussed by Landwehr et al.~\shortcite{landwehr2014model} but a much larger set of individuals (251 rather than 20).

Eye movement records for 251 individuals are obtained from an EyeLink II system with a 500-Hz sampling rate (SR Research, Ongoode, Ontario, Canada) while reading sentences from the \emph{Potsdam Sentence Corpus}~\cite{kliegl2006tracking}.
There are 144 sentences in the corpus, which we split into equally sized sets of training and test sentences.
Individuals read between 100 and 144 sentences, the training (testing) observations for one individual are the observations
on those sentences in the training (testing) set of sentences that the individual has read.
Results are averaged over 10 random train-test splits. 

We study the semiparametric model discussed in Section~\ref{sec:model} with MCMC inference as presented in Section~\ref{sec:inference} (denoted \ourModel).
We employ a squared exponential covariance function $\kappa(x,x') = \alpha \exp\left(-\frac{\|x-x'\|^2}{2 \sigma^2}\right)$, where the multiplicative constant $\alpha$ is tuned on the training
data and the bandwidth $\sigma$ is set to the average distance between points in the training data. The Beta and Dirichlet parameters $\lambda$ and $\rho$ are set to one
(Laplace smoothing), the prior $p(\bEta)$ for the Gamma parameters is uninformative. We use backoff-smoothing as discussed by Landwehr et al.~\shortcite{landwehr2014model}.
We initialize the sampler with the maximum-likelihood Gamma fit
and perform 10.000 sampling iterations, 5000 of which are burn-in iterations.
As a baseline, we study the model by Landwehr et al.~\shortcite{landwehr2014model} (\gammaModel)
and simplified versions proposed by them that only use saccade type and amplitude (\gammaModelTA)
or saccade type (\gammaModelT). 
We also study the weighted and 
unweighted version of the feature-based model of Holland \& Komogortsev~\shortcite{holland2012biometric} 
with a feature set adapted to the Potsdam Sentence Corpus data as described in Landwehr et al.~\shortcite{landwehr2014model}. 

\begin{table}[t]
\begin{tabular}{ l | c }
	Method & Accuracy\\
	\hline
	\ourModel & 0.9562 $\pm$ 0.0092 \\
	\gammaModel & 0.8319 $\pm$ 0.0218\\
	\gammaModelTA & 0.5964 $\pm$ 0.0262\\
	\gammaModelT & 0.2749 $\pm$ 0.0369 \\
	\hollandUnweighted & 0.6988 $\pm$ 0.0241\\
	\hollandWeighted & 0.4566 $\pm$ 0.0220\\
\end{tabular}
\caption{Multiclass identification accuracy $\pm$ standard error.}
\label{table:accresult}
\end{table}

We first study multiclass identification accuracy. All test observations of one particular individual constitute one test example; the task is to infer the individual that has generated 
these test observations. Multiclass identification accuracy is the fraction of cases in which the correct individual is identified. 
Table~\ref{table:accresult} shows multiclass identification accuracy for all methods. We observe that \ourModel outperforms \gammaModel, reducing the error by more than a factor of three.
Consistent with results reported in Landwehr et al.~\shortcite{landwehr2014model}, \hollandUnweighted is less accurate than \gammaModel, but more accurate than the simplified variants.
We next study how the amount of data available at test time---that is, the amount of time we can observe a reader before having to make a decision---influences accuracy. 
Figure~\ref{fig:difftestindvmulti} (left) shows identification accuracy as a function of the fraction of test data available, obtained by randomly removing a fraction of sentences from
the test set. We observe that identification accuracy steadily improves with more test observations for all methods. Figure~\ref{fig:difftestindvmulti} (right) shows identification
accuracy when varying the number $R$ of individuals that need to be distinguished. 
We randomly draw a subset of $R$ individuals from the set of 251 individuals,
and perform identification based on only these individuals. Results are averaged over 10 such random draws. As expected, accuracy improves if fewer individuals need to be distinguished.

\begin{figure}[t]
\centering
\includegraphics[width=0.99\linewidth,trim={0mm 0mm 0mm 0mm},clip]{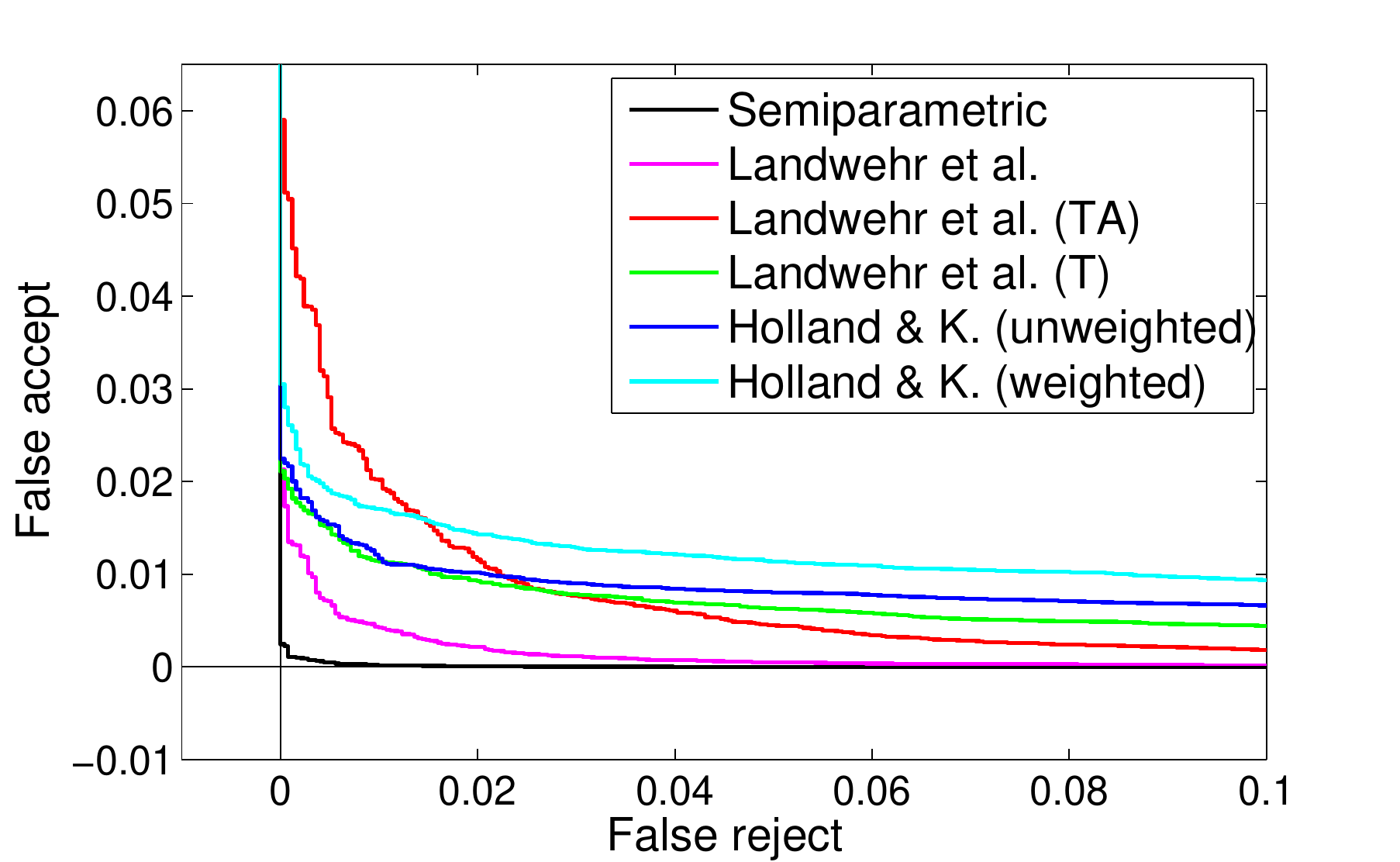}
\caption{False-accept over false-reject rate when varying~$\tau$.}
\label{fig:diffmethbin}
\end{figure}

\begin{figure}[t]
\centering
\includegraphics[width=0.99\linewidth,trim={0mm 0mm 0mm 0mm},clip]{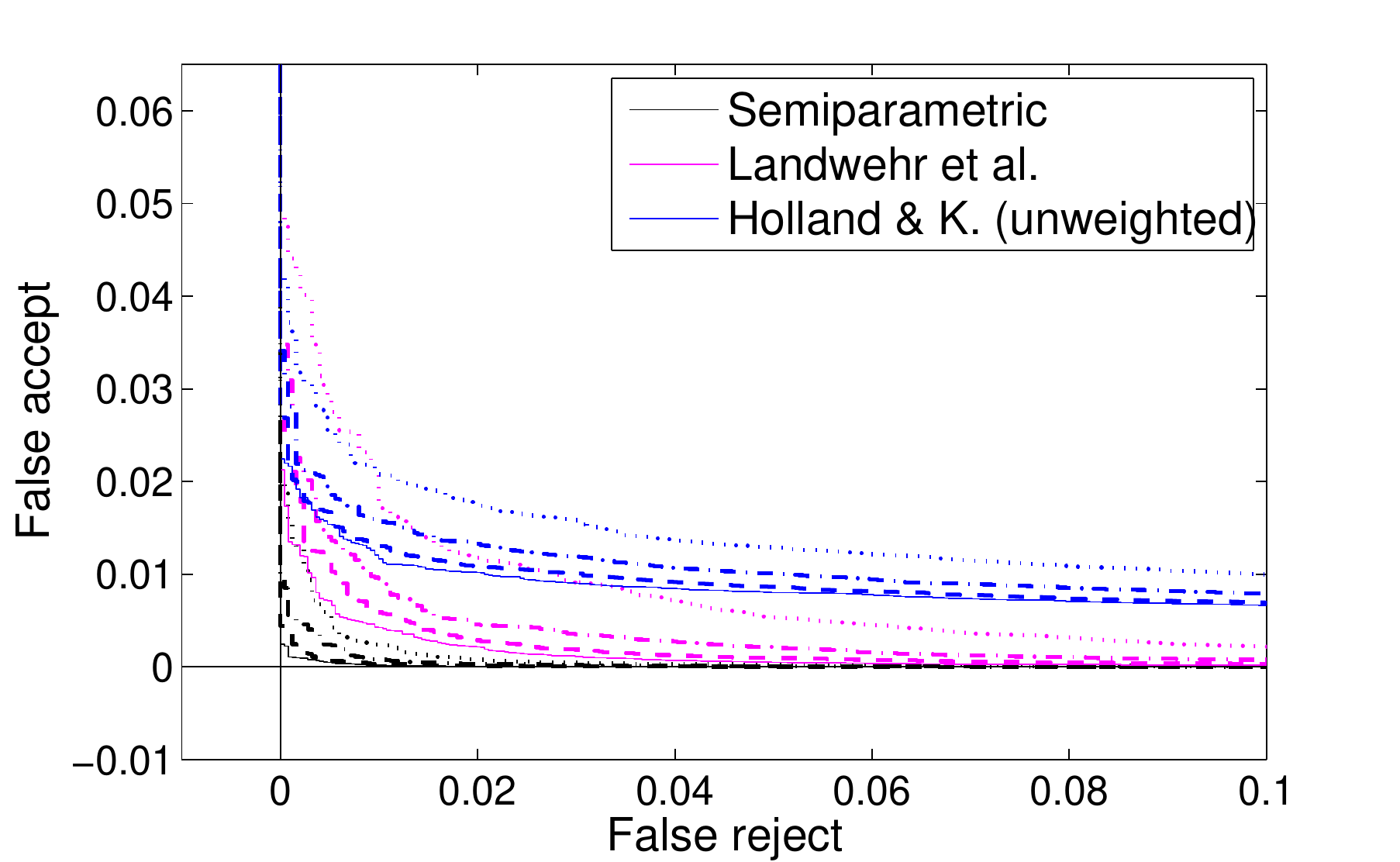}
\caption{
False-accept over false-reject rate when using 40\% (dotted), 60\% (dashed-dotted), 80\% (dashed), and 100\% (solid) of test observations, for selected subset of methods.}
\label{fig:difftestbin}
\end{figure}

\begin{table}[t]
\begin{tabular}{ l | c }
	Method & Area under curve \\
	\hline
	\ourModel & 0.0000098 \\
	\gammaModel & 0.0001743 \\
	\gammaModelTA &  0.0010371 \\
	\gammaModelT & 0.0017040 \\
	\hollandUnweighted & 0.0027853 \\
	\hollandWeighted & 0.0039978 \\
\end{tabular}
\caption{Area under the curve in binary classification setting.} 
\label{table:accresultbin}
\end{table}

We next study a binary setting in which for each individual and each set of test observations a decision has to be made whether or not the test observations have been generated by that 
individual. This setting more closely matches typical use cases for the deployment of a biometric system. 
Let $\bar{\mathcal{X}}$ denote the text being read at test time, and let $\bar{\mathcal{S}}$ denote the observed eye movement sequences. Our model infers for each reader $r\in\mathcal{R}$ 
the marginal likelihood $p(\bar{\mathcal{S}}|r,\bar{\mathcal{X}},\mathcal{X},\mathcal{S}^{(1:R)})$ of the eye movement observations under the reader-specific model (Equation~\ref{eq:goal_uniform_r}).
The binary decision is made by dividing this marginal likelihood by the average marginal likelihood assigned to the observations by all reader-specific models, and comparing the result to a threshold $\tau$.
Figure~\ref{fig:diffmethbin} shows the fraction of false accepts 
as a function of false rejects 
as the threshold $\tau$ is varied, averaged over all individuals. 
The \gammaModel model and variants also assign a reader-specific likelihood to  
novel test observations; we compute the same statistics again by normalizing the likelihood and comparing to a threshold $\tau$. Finally, \hollandUnweighted and \hollandWeighted compute a similarity measure
for each combination of individual and set of test observations, which we normalize and threshold analogously. We observe that \ourModel accomplishes a false-reject rate of below 1\% at virtually no false accepts;
\gammaModel and variants tend to perform better than \hollandUnweighted and \hollandWeighted. 
Table~\ref{table:accresultbin} shows the area under the curve for this experiment and the different methods. 
Figure~\ref{fig:difftestbin} shows the same experiment for different amounts of test observations and a selected subset of methods.

Training the joint model for all 251 individuals takes 46 hours on a single eight-core CPU; predicting the most likely individual to have generated a set of 72 test sentences takes less than 2 seconds.

\section{Conclusions}

We have studied the problem of identifying readers unobtrusively during reading of arbitrary text. 
For fitting reader-specific distributions, we employ a Bayesian semiparametric approach that infers densities under a Gaussian process prior
centered at the gamma family of distributions, striking a balance between robustness to sparse data and modeling flexibility. 
In an empirical study with 251 individuals, the model was shown to reduce identification error by more than
a factor of three compared to earlier approaches to reader identification proposed by Landwehr et al.~\shortcite{landwehr2014model} and Holland \& Komogortsev~\shortcite{holland2012biometric}.

\bibliographystyle{acl2012}
\bibliography{readerIdentification}

\end{document}